\pdfoutput=1

\documentclass[11pt]{article}

\PassOptionsToPackage{hyphens}{url}

\usepackage[preprint]{acl}

\usepackage{times}
\usepackage{latexsym}
\usepackage{lipsum}
\usepackage{stfloats}
\usepackage{amsmath}
\usepackage{booktabs}
\usepackage[T1]{fontenc}

\usepackage[utf8]{inputenc}

\usepackage{microtype}
\usepackage{inconsolata}

\usepackage{graphicx}

\title{Reward Modeling with Weak Supervision for Language Models}

\author{
  \textbf{Ben Hauptvogel\textsuperscript{1}},
  \textbf{Malte Ostendorff\textsuperscript{2}},
  \textbf{Georg Rehm\textsuperscript{2,3}},
  \textbf{Sebastian Möller\textsuperscript{1,3}}
\\
\\
  \textsuperscript{1}Technical University of Berlin~~
  \textsuperscript{2}Occiglot~~
  \textsuperscript{3}DFKI GmbH
\\
  \small{
    \textbf{Corresponding author:} \url{b.hauptvogel@tu-berlin.de}
  }
}

\begin{document}
\maketitle
\begin{abstract}

Recent advancements in large language models (LLMs) have led to their increased application across various tasks, with reinforcement learning from human feedback (RLHF) being a crucial part of their training to align responses with user intentions. In the RLHF process, a reward model is trained using responses preferences determined by human labelers or AI systems, which then refines the LLM through reinforcement learning. This work introduces weak supervision as a strategy to extend RLHF datasets and enhance reward model performance. Weak supervision employs noisy or imprecise data labeling, reducing reliance on expensive manually labeled data. By analyzing RLHF datasets to identify heuristics that correlate with response preference, we wrote simple labeling functions and then calibrated a label model to weakly annotate unlabeled data. Our evaluation show that while weak supervision significantly benefits smaller datasets by improving reward model performance, its effectiveness decreases with larger, originally labeled datasets. Additionally, using an LLM to generate and then weakly label responses offers a promising method for extending preference data. 

\end{abstract}

\section{Introduction}

Reinforcement learning from Human Feedback (RLHF) is a widely used method for aligning models to user intentions. This technique has been instrumental in improving large language models (LLM) to reflect human values and enhance usability, leading to large-scale adoption of conversational systems like ChatGPT \citep{openai2024gpt4} or BARD \citep{thoppilan2022lamda}.

The RLHF technique starts by sampling outputs from a model, which is either pre-trained or already supervised fine-tuned on demonstration data. Then, human annotators are tasked to label the outputs by ranking them from the least preferable to the most preferable. This labeled data is subsequently used to train a reward model, which calculates a reward value for a given response to a prompt. This is necessary for the reinforcement learning stage, in which a newly sampled model output is assigned this scalar reward. The model is then refined using an RL algorithm such as Proximal Policy Optimization (PPO) \citep{ouyang2022training, schulman2017proximal}. During this process, the collection of high-quality human feedback data presents a significant challenge since it remains an expensive task \citep{casper2023open}.

An alternative to relying on labeled datasets is the approach of weak supervision. Weak supervision is a machine learning technique that deviates from relying solely on manually labeled data. Instead, models are trained using noisy and inaccurate labels. A popular approach for implementing weak supervision involves the use of labeling functions. These are defined using programmatic rules and heuristics about the data and contain uncertain accuracies and correlations. Snorkel is a solution that denoises the labeling functions to create a weak supervision signal, without the need to specify weights \citep{ratner2020snorkel}.

Building on the advancements of model alignment techniques, this work focuses on the effectiveness of applying weak supervision to extend RLHF datasets. We aim to investigate whether annotation based on simple heuristics that model preference can enhance reward model performance. To ensure reproducibility we make all our source code and datasets publicly available on Github\footnote{\url{https://github.com/DFKI-NLP/weak-supervision-rlhf}}.

\section{Related Work}

Several works aim to remove the human labor in the annotation process from the RLHF pipeline. \citet{lee2023rlaif} use an off-the-shelf LLM to annotate preference samples instead of relying on human labeling. Their research concentrated on summarization tasks and found that reinforcement learning with AI feedback can achieve similar performance as RLHF. \citet{sun2023salmon} extend this approach by introducing guidelines for a reward model to address the reward hacking problem, in which a model tries to bypass the true objective by finding unintended ways to maximize its reward. 
\citet{kim2023aligning} align an LLM with synthetic feedback by employing heuristics based on a set of assumptions, which include the belief that larger models outperform smaller ones and that using more examples (shots) is preferable to using fewer. Samples generated using these characteristics were ranked higher in the preference dataset. 

Other studies explore methods to approximate human preference. \citet{bukharin2023deep} use domain knowledge to rank reward factors, creating a hierarchical decision tree to weakly annotate samples. In contrast to prior approaches, this work employs weak supervision rather than a decision tree, combining different reward factors to annotate samples based on a weak preference signal. Some suspect that output length plays a significant role in optimizing reward models. \citet{singhal2023long} explore the correlation between output length and reward, finding that the majority of reward improvements are due to length increases. Our work investigates length and other reward factors, involving the analysis analysis of multiple RLHF datasets to assess correlations between factors and corresponding rewards.


\section{Methodology}

\begin{figure*}[t]
  \includegraphics[width=\linewidth]{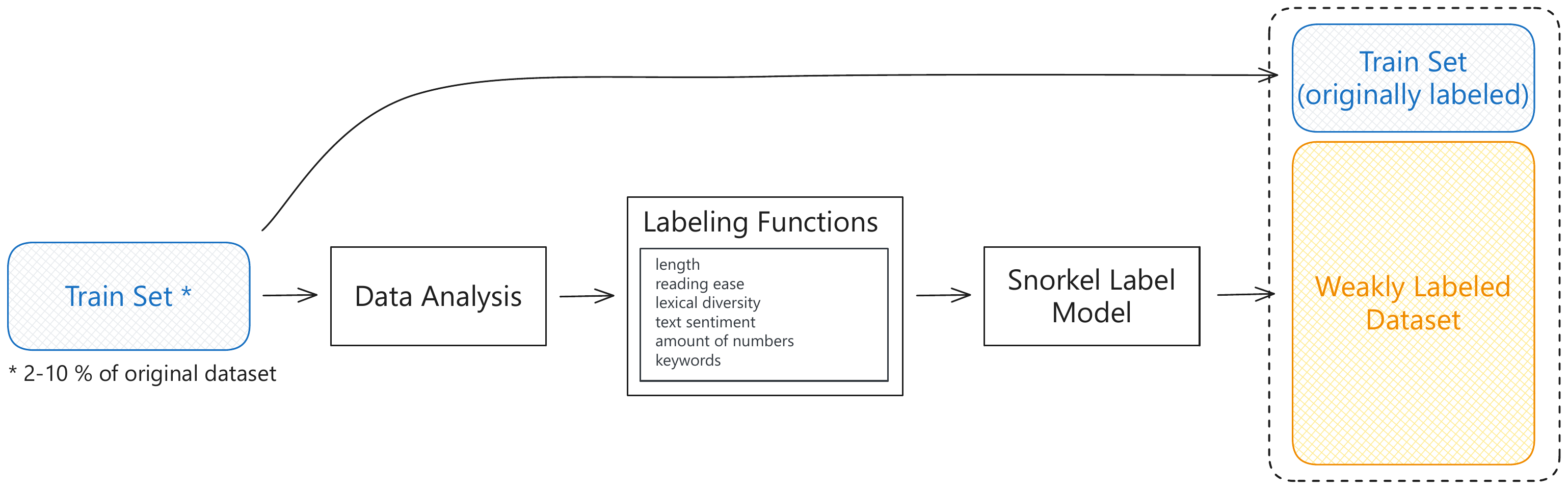}
  \caption{Extending RLHF datasets with weak supervision in a three-step pipeline: conducting data analysis, writing labeling functions, applying a label model to create a new weakly labeled dataset.}
  \label{fig:overview}
\end{figure*}

Our approach employs weak supervision to extend reinforcement learning from human feedback (RLHF) datasets. We start by analyzing parts of these datasets to identify heuristics that correlate with user preferences, which we use to develop labeling functions. These functions are combined using a linear label model which is able to weakly annotate unlabeled data. The resulting dataset with noisy preference data is combined with the originally labeled data to train a reward model.

\subsection{Datasets}
\label{sec:datasets}

We conducted experiments using four different preference datasets. For two of these datasets, human labelers were tasked to determine response preference, whereas for the remaining two, a LLM was employed to decide the preferred response.

The HH-RLHF dataset\footnote{\url{https://huggingface.co/datasets/Anthropic/hh-rlhf}} from Anthropic AI was constructed through a series of interactions between crowdworkers and LLMs in dialogue settings \citep{bai2022training}. At each node of the dialogue, the crowdworkers were presented with two model generated responses. They selected the response that was more helpful and less harmful. This process yielded a dataset containing about 169 thousand chosen-rejected response pairs.

The \textit{mt\_bench\_human\_judgements} dataset\footnote{\url{https://huggingface.co/datasets/lmsys/mt_bench_human_judgments}}, referred to as MT-BENCH for simplicity, is a human-annotated preference dataset with responses generated by six LLMs including GPT-4, GPT-3.5, and others \citep{zheng2023judging}. Graduate students with expertise in relevant subjects primarily annotated the responses to assess the alignment between human preferences and an LLM judge (GPT-4). Featuring about 3,300 samples, this dataset is considerably smaller than the HH-RLHF dataset.

The \textit{ultrafeedback-binarized} (UB) dataset\footnote{\url{https://huggingface.co/datasets/HuggingFaceH4/ultrafeedback_binarized}} employs an LLM, specifically OpenAI's GPT-4, for response ranking across 64 thousand prompts collected from various sources, generating four responses per prompt and annotating each based on instruction-following, truthfulness, honesty, or helpfulness, and an overall score for preference \citep{cui2023ultrafeedback, openai2024gpt4}. However, due to inconsistencies in the overall scores, researchers at Argilla recalculated using the mean of the objective ratings to form the ultrafeedback-binarized-preferences (UBP) dataset\footnote{\url{https://huggingface.co/datasets/argilla/ultrafeedback-binarized-preferences}}. In this dataset, they used the highest-rated response as the chosen option and randomly selected one of the remaining responses as the rejected counterpart for pairwise comparisons.

Ten percent of each dataset was held-out as an evaluation set, excluded from the processing pipeline. The remaining data was further divided into a baseline training set, comprising between 1 and 10\% of the total dataset, and a weakly supervised set. From this latter part, original preference labels were removed and replaced with newly applied weak labels.

An exception to this is the MT-BENCH dataset. Due to its small size, 30\% is used as the evaluation set, with the remaining 70\% designated as the baseline training set.
Since we did not use any data of it for weak supervision, we adopted a different strategy for the weakly labeled dataset by generating new unlabeled data consisting of a prompt and two responses. First, we compiled prompts from various datasets including HH-RLHF, OpenAssistant, alpaca, and Synthetic Instruct GPTj Pairwise. We then generated responses using LlaMa-2-7b and Vicuna-7b-v1.5 LLMs, ensuring comparability by choosing models with the same parameter size. In total, we generated around 24,200 prompt-response-response triplets, which were uploaded to Hugging Face.

\subsection{Heuristics}

The selection of heuristics that potentially correlate with human or AI preference was primarily driven by theoretical considerations, an intuitive understanding of response dynamics, and insights from existing literature on RLHF reward factors. Text length was the first feature we investigated, since most RLHF datasets show a strong correlation between the response length and its preference \citep{singhal2023long}.

Next, we applied a formula to assess the readability of a text using the Flesch Reading Ease, which calculates readability based on the total number of words, sentences, and syllables \citep{flesh1948readability}.
The Flesch Reading Ease score indicates how easy or difficult a text is to read. Lower scores indicate the text is more challenging to read. The highest possible score is 121.22, which represents the easiest readability. Typically, scores for most texts range from 0 to 100.

We analyzed the lexical diversity in the datasets, which is a calculated measure of vocabulary richness within a text. Lexical diversity indicates the variety of different words used, relative to the total word count. We employed the Type-Token Ratio for this analysis, which calculates lexical diversity by dividing the number of unique words by the total number of words in the text.

Next, we counted the amount of numbers in each response to determine if there is a relationship between the quantity of numbers and preference.

Additionally, we conducted sentiment analysis on the response texts. Sentiment analysis uses computational methods to determine the emotional tone of a text, categorizing it as positive, negative, or neutral. For this purpose, we used the Valence Aware Dictionary and Sentiment Reasoner (VADER), a lexicon and rule-based tool for sentiment analysis \citep{hutto2014vader}. Using VADER, we assessed the sentiment polarity. Sentiment polarity identifies the emotional direction of the content, showing whether the text conveys a positive, negative, or neutral message.

We used an external LLM to generate regular expressions that are potentially more common in either chosen or rejected responses. We tracked how frequently these expressions appeared in each type of response. If a regular expression appears significantly more often in a chosen response than in a rejected response, it can be useful to integrate into a labeling function.

Finally, we also used keywords to label responses. For this purpose, we collected multiple lists of harmful or offensive keywords from the Internet. The presence of these keywords in a text often indicates that the response could be more harmful or offensive. We validated this pattern within our datasets.

\subsection{Data Analysis}

For each heuristic that potentially influences the reward, we conducted a detailed data analysis before developing labeling functions based on those findings. This data analysis involves determining whether a correlation exists between the heuristic and preference, and determining if its relevance is confined to a specific range of values.

The data analysis was conducted on the 10 \% train split of each dataset. We examined numerical features, such as response length or amount of numbers, by analyzing the average values for both chosen and rejected responses. An independent t-test on these averages determined if the differences were statistically significant. Some of the resulting p-values were above 0.05, indicating that the difference is not statistically significant, but we still implemented labeling functions for those heuristics. They can still provide a valuable weak supervision signal since the label model will weigh the labeling functions based on their accuracy and correlations. The Snorkel label model is robust to noise, so providing additional context, even if not always precise can help differentiate edge cases \citep{ratner2018training}.

\begin{table*}[t]
  \centering
  \begin{tabular}{lrr|rr|rr|rr}
    \toprule
     & \multicolumn{2}{c|}{\textbf{HH-RLHF}} & \multicolumn{2}{|c|}{\textbf{UB}} & \multicolumn{2}{|c|}{\textbf{UBP}} & \multicolumn{2}{|c}{\textbf{MT-BENCH}} \\
    
    Feature & stat & p-value & stat & p-value & stat & p-value & stat & p-value \\
    \midrule
    Text Length & 4.12 & < 0.01 & 9.38 & < 0.01 & 18.12 & < 0.01 & 5.92 & < 0.01 \\
    Reading Ease & -4.15 & < 0.01 & -1.60 & 0.11 & -4.11 & < 0.01 & -1.96 & 0.05 \\
    Lexical Diversity & -5.60 & < 0.01 & -1.95 & 0.05 & -9.89 & < 0.01 & -7.28 & < 0.01 \\
    Amount of Numbers & 1.49 & 0.14 & 5.53 & < 0.01 & 10.33 & < 0.01 & 3.11 & < 0.01 \\
    Sentiment Polarity & 1.49 & < 0.01 & 5.53 & 0.84 & 10.33 & < 0.01 & 3.11 & < 0.01 \\
    \bottomrule
  \end{tabular}
  \caption{Results of the independent t-test for numerical features of RLHF datasets.
  }
  \label{tab:t-test_results}
\end{table*}

We found a clear correlation that longer responses are consistently more likely to be chosen over shorter ones. The average length of chosen responses is longer than that of rejected responses across all four datasets. The t-test results confirm that this difference is statistically significant, with all four p-values well below the 0.05 threshold, as shown in Table \ref{tab:t-test_results}.

The average reading ease score for rejected responses is higher than for chosen responses across all four datasets, indicating that preferred responses are generally more challenging to read. The t-test confirms the statistical significance of this trend for the HH-RLHF, MT-BENCH, and UB datasets, with p-values below 0.05. However, for the UB dataset, the p-value of 0.11 is not statistically significant. Despite this, we will continue to incorporate reading ease into the labeling functions for all datasets and assess their effectiveness.

The average lexical diversity is lower in chosen responses than in rejected responses. The p-value from the independent t-tests confirms that this observation is statistically significant for all datasets. Consequently, our labeling function for lexical diversity favors responses with lower lexical diversity.

For the HH-RLHF datasets the chosen responses generally include more numbers on average in all datasets, but the difference is not statistically significant. In contrast, for the other datasets, the chosen responses contain a statistically significant higher amount of numbers compared to rejected responses. We developed a labeling function that favors responses containing more numbers.

Finally, the sentiment polarity, as calculated by VADER, is generally higher for chosen responses compared to rejected responses across all four datasets. A t-test validates these findings, confirming that the mean difference in sentiment polarity is statistically significant for all datasets except for the UB dataset. Consequently, we have developed labeling functions that favor responses with higher sentiment polarity.

We conducted further analysis on these numerical features to determine if the observed correlations are confined to specific ranges. For the non-numerical features, lists of regular expressions and keywords, a different approach was taken. GPT-4 was used to generate regular expressions that could influence response preferences. Prompts were formulated to produce regular expressions common in chosen or rejected responses. For example, rejected responses might include expressions of uncertainty, while chosen responses might include pros and cons or specific examples.

We counted how frequently these regular expressions appeared in both chosen and rejected responses. When a regular expression demonstrated a statistically significant variance in occurrence between chosen and rejected responses and occurred frequently in general, it was integrated into the labeling function. We established specific thresholds for the minimum occurrence ration and overall frequency required. A regular expression that appeared with at least a 10\% higher frequency in either chosen or rejected responses was adopted for that respective group in the labeling function. The resulting labeling function consists of two lists, positive and negative regular expressions. When comparing two responses, it outputs the response that contains more of the positive and fewer of the negative expressions. Since the occurrences of regular expressions vary across datasets, the lists of positive and negative expressions are different for each dataset.


Very similar to using regular expressions, we also used lists of negative keywords for labeling functions. We collected lists of words from the internet that we believe are more likely to appear in bad responses. Three distinct lists were used in the analysis: one containing offensive words\footnote{\url{https://github.com/LDNOOBW/List-of-Dirty-Naughty-Obscene-and-Otherwise-Bad-Words/}}, which are normally used for moderating user-generated content, one containing harmful words, and a large list of negatively connotated words\footnote{\url{http://www.bannedwordlist.com/}}, primarily consisting of obscene or vulgar terms, which we will refer to as ``bad'' words for simplicity.

\begin{table}
  \centering
  \begin{tabular}{lcc}
    \toprule
    \multicolumn{3}{c}{Occurrences in \textbf{HH-RLHF}} \\
    & \textbf{Chosen} & \textbf{Rejected} \\
    \midrule
    Offensive Words & 139 & 221 \\
``Bad'' Words & 285 & 402 \\
Harmful Words & 616 & 779 \\
    \midrule
    \multicolumn{3}{c}{Occurrences in \textbf{UB}} \\
    & \textbf{Chosen} & \textbf{Rejected} \\
    \midrule
    Offensive Words & 20 & 23 \\
``Bad'' Words & 82 & 75 \\
Harmful Words & 235 & 200 \\
    \midrule
    \multicolumn{3}{c}{Occurrences in \textbf{UBP}} \\
    & \textbf{Chosen} & \textbf{Rejected} \\
    \hline
    Offensive Words & 41 & 38 \\
``Bad'' Words & 101 & 70 \\
Harmful Words & 317 & 238 \\
    \midrule
    \multicolumn{3}{c}{Occurrences in \textbf{MT-BENCH}} \\
    & \textbf{Chosen} & \textbf{Rejected} \\
    \hline
    Offensive Words & 0 & 5 \\
``Bad'' Words & 9 & 3 \\
Harmful Words & 17 & 9 \\
    \bottomrule
  \end{tabular}
\caption{Occurrences of words from three keyword lists in chosen and rejected responses across datasets.}
\label{tab:keyword-occurrences}
\end{table}

Table \ref{tab:keyword-occurrences} shows a clear difference between the human-annotated HH-RLHF dataset and the other datasets. In the HH-RLHF dataset, the words of all three keyword lists are more commonly found in rejected responses, which aligns with the dataset's goals to be helpful and harmless. In the AI-annotated UB and UBP datasets, the trend is reversed, with chosen responses containing offensive, harmful, or bad words more frequently. However, it is important to highlight that only a small number of responses contained words from these lists. In the UB dataset for example, among the 4,828 chosen and rejected responses in the train set, there were fewer than 450 harmful words, fewer than 150 ``bad'' words, and fewer than 50 offensive words (similar in the UBP dataset). Even fewer words were found in the MT-BENCH set, which is understandable given its smaller size of just 898 chosen and rejected responses in the set we analyzed.

Therefore, we decided not to write labeling functions based on these keyword findings for the UB, UBP, and MT-BENCH datasets, as we do not believe this pattern -- more negative words in preferred responses -- will generalize well to new data. We prefer not to base our labeling functions on the prevalence of more negative words. However, for the HH-RLHF dataset, we created a labeling function for each list to count these keywords and favor the response with fewer of them.

\subsection{Labeling Functions}

Based on our data analysis results, we developed labeling functions. These concise functions take two responses as input and select a preferred response according to a defined, simple heuristic or abstain from making a decision.

The developed labeling functions were applied to each train set. We further validated the efficacy of the labeling functions using two primary metrics, coverage and accuracy. The (empirical) accuracy reflects how often the labeling function correctly identifies the actual preferred response. Coverage indicates how frequently the labeling functions make a decision instead of abstaining. 

Labeling functions abstain from making decision either due to identical heuristic values between responses or due to predefined cutoff points. These cutoff points are based on the data analysis, which identified ranges where the effects of heuristics are stronger or weaker. Beyond those cutoff points the labeling functions abstain, reducing their coverage but potentially enhancing accuracy. While a grid search could be used to determine these thresholds on each train set for optimal coverage and accuracy, our primary goal with these labeling functions is not solely to optimize performance on the 10\% train set. We aim to ensure they generalize well on the remainder of the dataset or unseen data.

\begin{table}[h]
  \centering
  \begin{tabular}{lrr}
    \toprule
    \textbf{Labeling function} & \textbf{Coverage} & \textbf{Accuracy} \\
    \midrule
    Length              & 88.54\% & 52.36\% \\
Reading ease        & 74.50\% & 52.74\% \\
Lexical diversity   & 50.81\% & 53.65\% \\
Sentiment polarity  & 83.68\% & 52.39\% \\
Amount of numbers   & 6.99\%  & 53.31\% \\
Regular Expressions & 27.93\% & 54.40\% \\
Offensive keywords  & 1.31\%  & 60.00\% \\
Harmful keywords    & 4.42\%  & 57.75\% \\
Bad Keywords        & 1.89\%  & 57.30\% \\
    \bottomrule
  \end{tabular}
  \caption{Labeling functions analysis on train set (10\% of the HH-RLHF dataset).}
\label{tab:lfs-hh_rlhf}
\end{table}

Table \ref{tab:lfs-hh_rlhf} shows the labeling function for the HH-RLHF dataset. Each labeling function achieves an accuracy exceeding 50\% on the train set. However, none surpass 60\%, indicating that these simple heuristics do not a highly accurate reflection of the human preference represented in this dataset. The coverage of labeling functions varies significantly. For numerical values, coverage depends on the established thresholds. Coverage for keyword lists is expectedly low due to the rarity of negative words in model-generated responses. Similarly, differences in the amount of numbers between responses are rare.

\begin{table}[h]
  \centering
  \begin{tabular}{lrr}
    \toprule
    \textbf{Labeling function} & \textbf{Coverage} & \textbf{Accuracy} \\
    \midrule
    Length                 & 95.32\% & 69.97\% \\
Reading ease           & 69.26\% & 60.45\% \\
Lexical diversity      & 62.13\% & 61.69\% \\
Sentiment polarity     & 69.93\% & 59.39\% \\
Amount of numbers      & 63.47\% & 63.50\% \\
Regular Expressions    & 30.62\% & 58.54\% \\
    \bottomrule
  \end{tabular}
  \caption{Labeling functions analysis on train set (MT-BENCH dataset).}
\label{tab:lfs-mt_bench}
\end{table}

Table \ref{tab:lfs-mt_bench} shows the labeling functions used for the MT-BENCH dataset. The accuracies of the labeling functions are notably higher than those for the other dataset. For instance, the labeling function for text length achieves an empirical accuracy of almost 70\%, while the same labeling function applied to the HH-RLHF dataset achieves an accuracy of about 52 \%. It is important to note, however, that the MT-BENCH dataset is considerably smaller than the HH-RLHF dataset.

\begin{table}[h]
  \centering
  \begin{tabular}{lrr}
    \toprule
    \textbf{Labeling function} & \textbf{Coverage} & \textbf{Accuracy} \\
    \midrule
Length                 & 93.61\% & 56.99\% \\
Reading ease           & 68.21\% & 55.30\% \\
Lexical diversity      & 52.19\% & 53.94\% \\
Sentiment polarity     & 65.13\% & 55.13\% \\
Amount of numbers      & 62.75\% & 61.43\% \\
Regular Expressions    & 32.02\% & 57.46\% \\
    \bottomrule
  \end{tabular}
\caption{Labeling functions analysis on train set (10\% of the UB dataset).}
\label{tab:lfs-UB}
\end{table}

Table \ref{tab:lfs-UB} shows the labeling functions applied to the UB dataset, and Table \ref{tab:lfs-UBP} presents those applied to the UBP dataset. Both datasets exhibit similar coverages, but the accuracies are notably higher for the UBP dataset compared to the UB dataset.

\begin{table}[h]
  \centering
  \begin{tabular}{lrr}
    \toprule
    \textbf{Labeling function} & \textbf{Coverage} & \textbf{Accuracy} \\
    \midrule
    Length                 & 95.06\% & 67.43\% \\
Reading ease           & 68.98\% & 57.80\% \\
Lexical diversity      & 52.88\% & 63.90\% \\
Sentiment polarity     & 64.60\% & 55.85\% \\
Amount of numbers      & 50.61\% & 71.70\% \\
Regular expressions    & 29.11\% & 60.77\% \\
    \bottomrule
  \end{tabular}
\caption{Labeling functions analysis on train set (10\% of the UBP dataset).}
\label{tab:lfs-UBP}
\end{table}

\subsection{Label Model}

We fitted the Snorkel label model using the listed labeling functions and the train set for calibration. The model was fitted over 100 epochs with an L2 regularization of 0.5 and using an Adam optimizer. Once calibrated, the label model combines the labeling functions and can provide a probability classification for any given input. In the context of preference classification, it predicts the probability of one response being preferred over another based on the heuristics.

\begin{table}[h]
  \centering
  \begin{tabular}{lp{2cm}p{2cm}}
    \toprule
    \textbf{Dataset} & \textbf{Accuracy on train set} & \textbf{Accuracy on weak set} \\
    \midrule
    HH-RLHF & 53.17\% & 52.97\% \\
    MT-BENCH & 67.82\% & \textit{N.A.} \\
    UB & 57.42\% & 56.56\% \\
    UBP & 66.03\% & 64.45\% \\
    \bottomrule
  \end{tabular}
\caption{Label Model classification accuracy on train set and weakly labeled set. The weakly labeled set for the MT-BENCH dataset is not part of the original set, as explained in section \ref{sec:datasets}. Due to the absence of gold labels, it is not possible to compute the label model accuracy.}
\label{tab:label-model-accuracies}
\end{table}

We applied the label model to the remainder of each dataset, now referred to as the weakly labeled dataset and assessed the accuracy of the label model by comparing the label model outputs to the original labels. Table \ref{tab:label-model-accuracies} shows the achieved classification accuracies on the train sets and the weakly labeled sets. The accuracies on the weakly labeled sets are very similar, only slightly worse, compared to the train sets.

\subsection{Confidence Thresholds}

The label model we calibrated generates a prediction probability. for each class. Samples with a probability below 0.5 are classified as 0, and those above 0.5 as 1. In our context, a 0 indicates a preference for response 0, and conversely for a 1. This probability reflects the model's confidence in its prediction. We converted the prediction probability into a confidence value for each sample.

\begin{equation}
  \label{eq:confidence-threshold}
  \text{confidence}=\begin{cases}
    P & \text{if } P\geq 0.5\\
    1-P,              & P<0.5
\end{cases}
\end{equation}

where $P$ = prediction probability.

To improve the accuracy of our labeling, we can implement a confidence threshold. We specify a particular threshold value and exclude any samples with confidence levels below this value. This technique can increase average accuracy, but it comes with the trade-off of reducing the number of weakly labeled samples. We conducted experiments with different confidence thresholds to assess their impact on the reward model performance.

\subsection{Experiments} 

After applying weak supervision, we obtained the weakly labeled datasets, some of which were filtered using various confidence thresholds, alongside the train set used for labeling function data analysis and label model calibration. We trained a baseline reward model using the train set. For our experiments, we combined the various weakly labeled datasets with the corresponding train set to train a new reward model. We conducted the training of the reward model on the DFKI High-Performance-Compute cluster over two training epochs, using a learning rate of 8e-6 and a residual dropout rate of 0.01. Additionally, we used float32 as the datatype. As a base model architecture, we utilized DeBERTa V3. 

After training a reward model, either on a baseline train set or a weakly supervised dataset, it was evaluated using the corresponding evaluation set. During this phase, the model performed inference to classify the samples within the evaluation set. Classification accuracy was determined by the model's ability to correctly identify the preferred response, with incorrect classifications reflecting failures. We primarily used the F1 score to quantify accuracy because it balances precision and recall, making it ideal for our analysis.

\section{Results}

We evaluated the impact of different baseline training set sizes to determine how the availability of more or fewer originally labeled data affects performance. The results are illustrated in plots that show the F1 performance relative to the number of weakly labeled samples used. Each plot's x-axis shows the amount of weakly annotated data added to the baseline. These samples were selected based on falling below a specified confidence threshold, ensuring they represent the N highest-confidence samples. Detailed results for all datasets, including all F1 scores, the numbers of baseline and weakly labeled samples used, and confidence thresholds, can be found in Appendix~\ref{sec:experiment-results}.

\subsection{HH-RLHF}

\begin{figure*}[t]
  \includegraphics[width=0.48\linewidth]{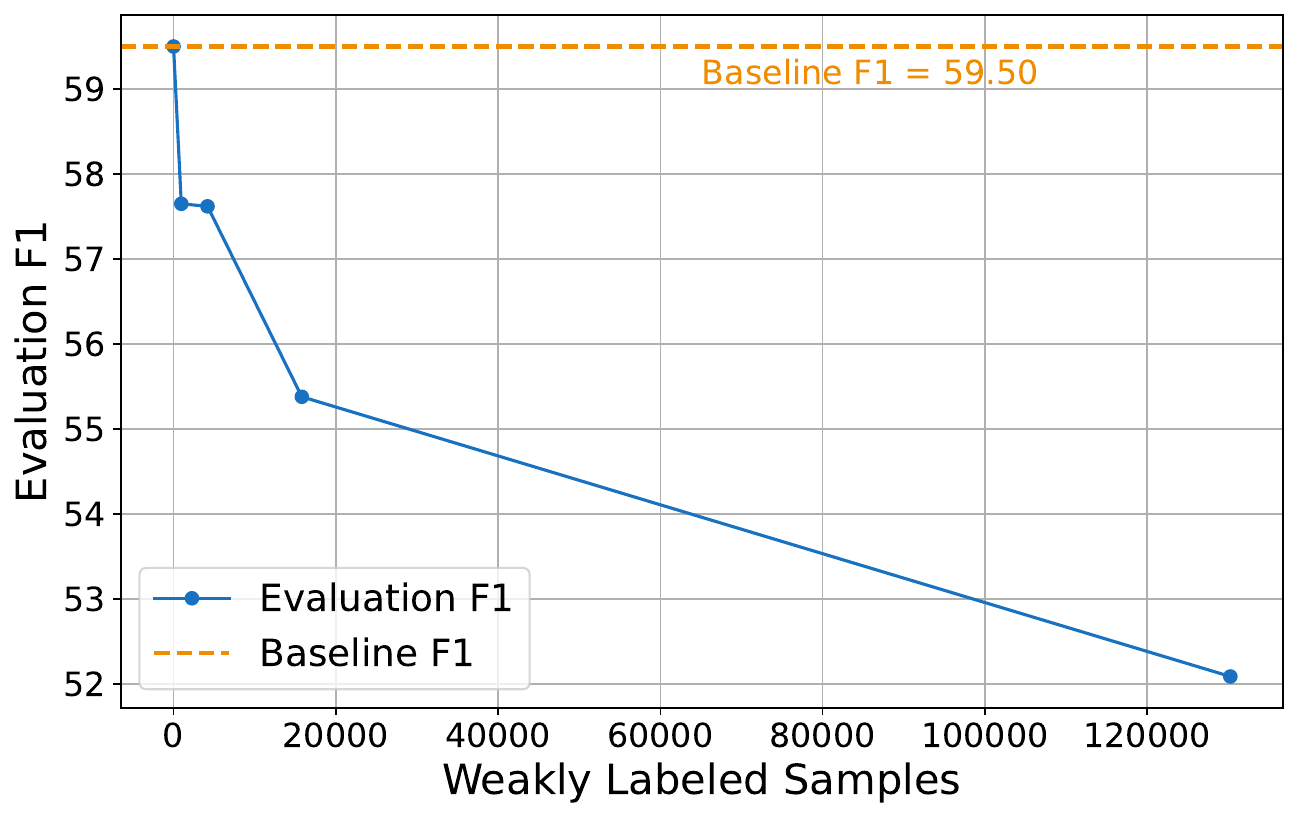} \hfill
  \includegraphics[width=0.48\linewidth]{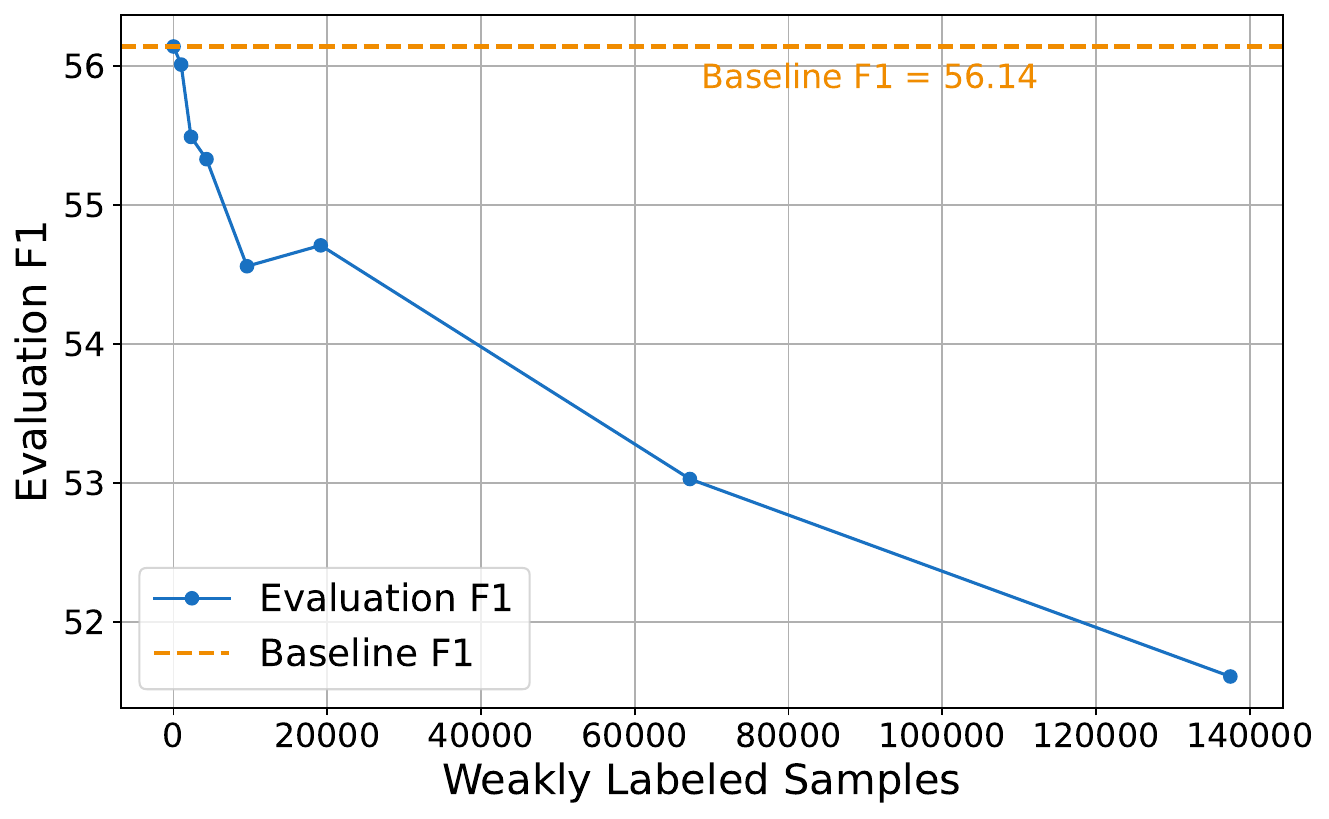}
  \caption {Evaluation for HH-RLHF using 10\% (left) and 5\% (right) as a baseline train set.}
  \label{fig:results_hh-rlhf_5_10}
\end{figure*}

Figure \ref{fig:results_hh-rlhf_5_10} demonstrates that there is no improvement in extending the train set with our weak supervision pipeline, using a baseline train set size of 10\% (14,472 samples) or 5\% (7,236 samples). The baseline F1 scores of 59.5\% and 56.14\% are not particularly high, especially compared to the performance of models trained on the other datasets.

\begin{figure*}[t]
  \includegraphics[width=0.48\linewidth]{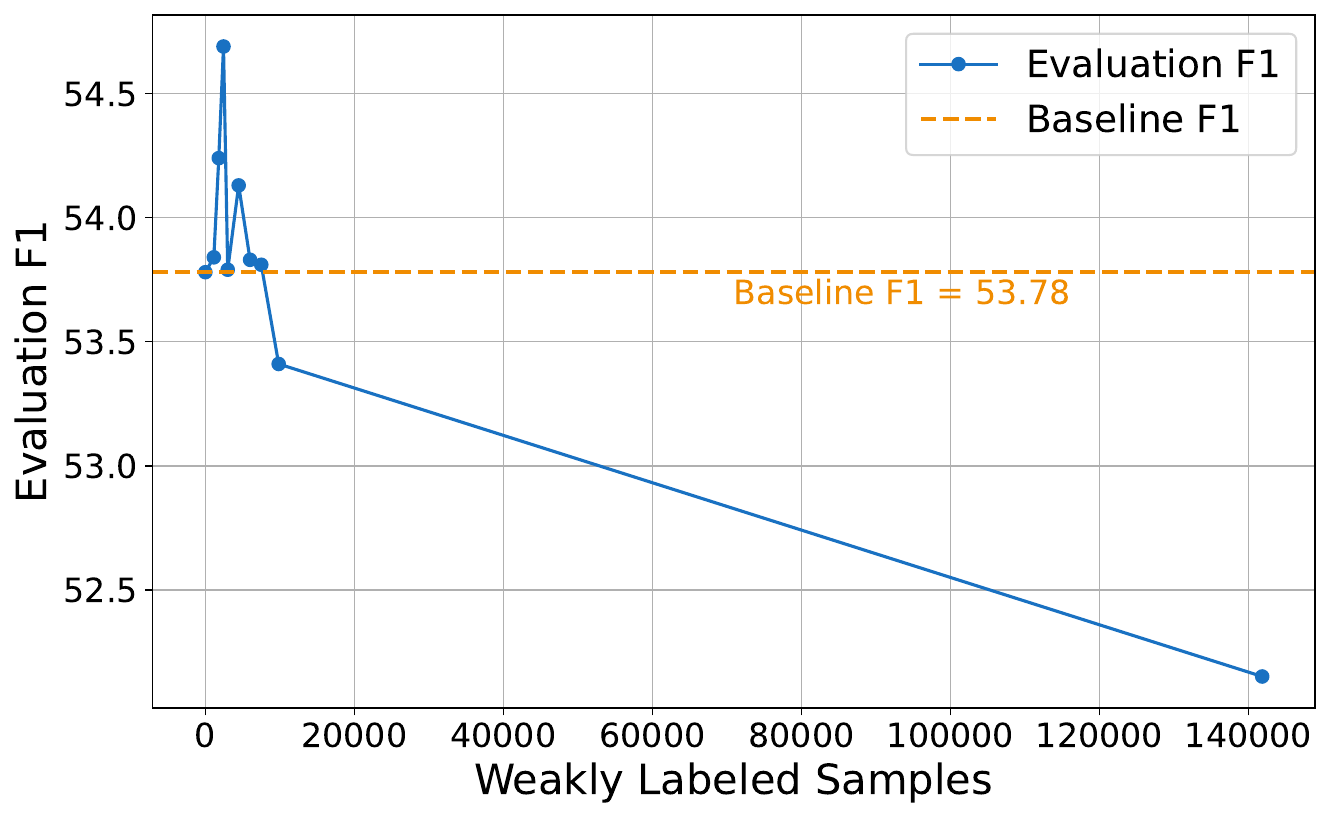} \hfill
  \includegraphics[width=0.48\linewidth]{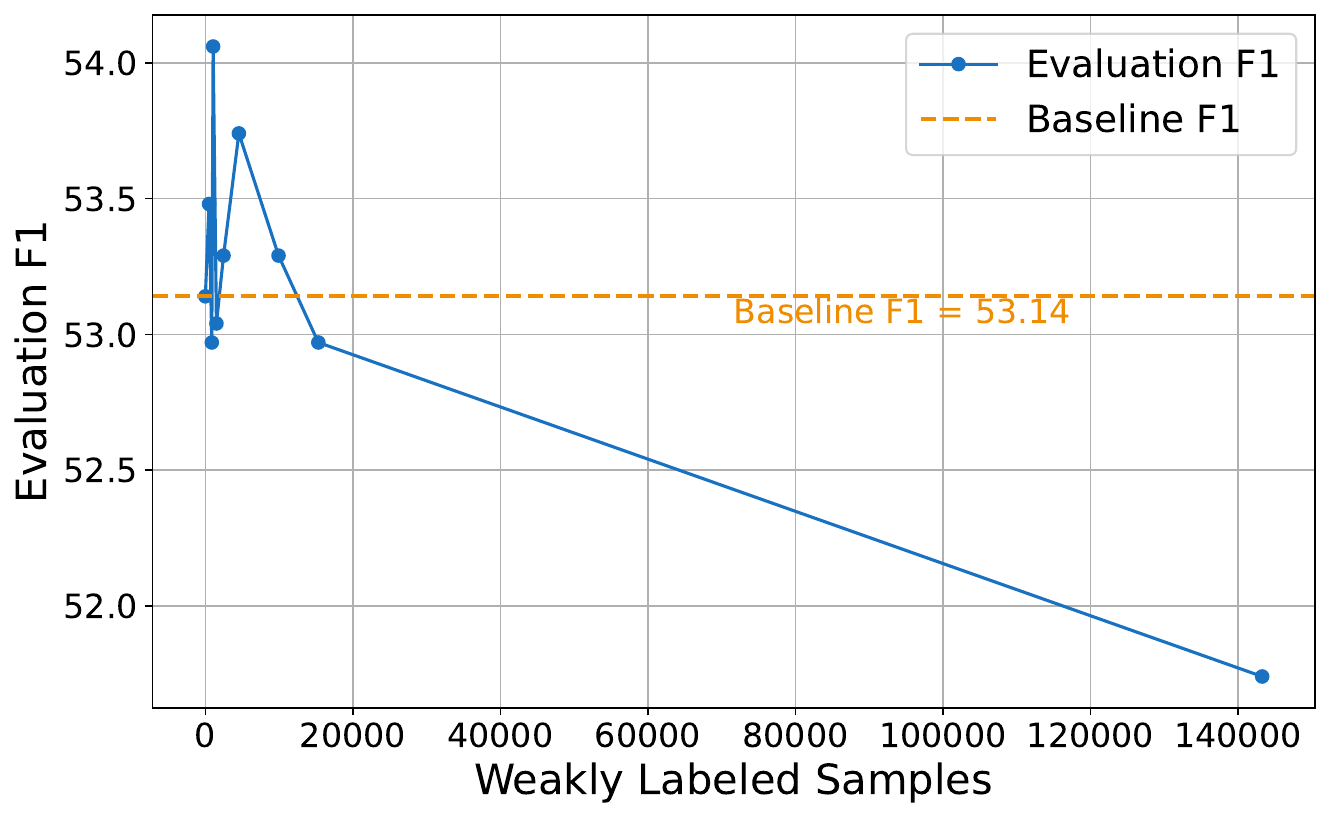}
  \caption {Evaluation for HH-RLHF using 2\% (left) and 1\% (right) as a baseline train set.}
  \label{fig:results_hh-rlhf_2_1}
\end{figure*}

Using a smaller baseline of 2\% or 1\%, weak supervision shows a performance improvement over their respective baseline scores. The performance for the 2\% baseline set (2,895 samples) reaches a peak at an F1 score of 54.69\%, compared to the baseline F1 of 53.78\%. While not a substantial increase, this results is notably different from those obtained with 5\% and 10\% baselines sets. 

Given the larger size of the HH-RLHF dataset, we also implemented a 1\% baseline set. The baseline F1 performance of 53.14\% was improved when adding weakly annotated samples. The best result was achieved when adding 1,051 weakly annotated samples to the 1,448 originally labeled samples, which resulted in an F1 score of 54.06\%. However, performance declined with the use of more weakly annotated samples. Additionally, results are more volatile with a smaller number of training samples, as each significantly influences the training process. This volatility is evident with the spikes and fluctuations in Figure \ref{fig:results_hh-rlhf_2_1} with a 1\% baseline.

\subsection{UB}

\begin{figure*}[t]
\centering
  \includegraphics[width=0.48\linewidth]{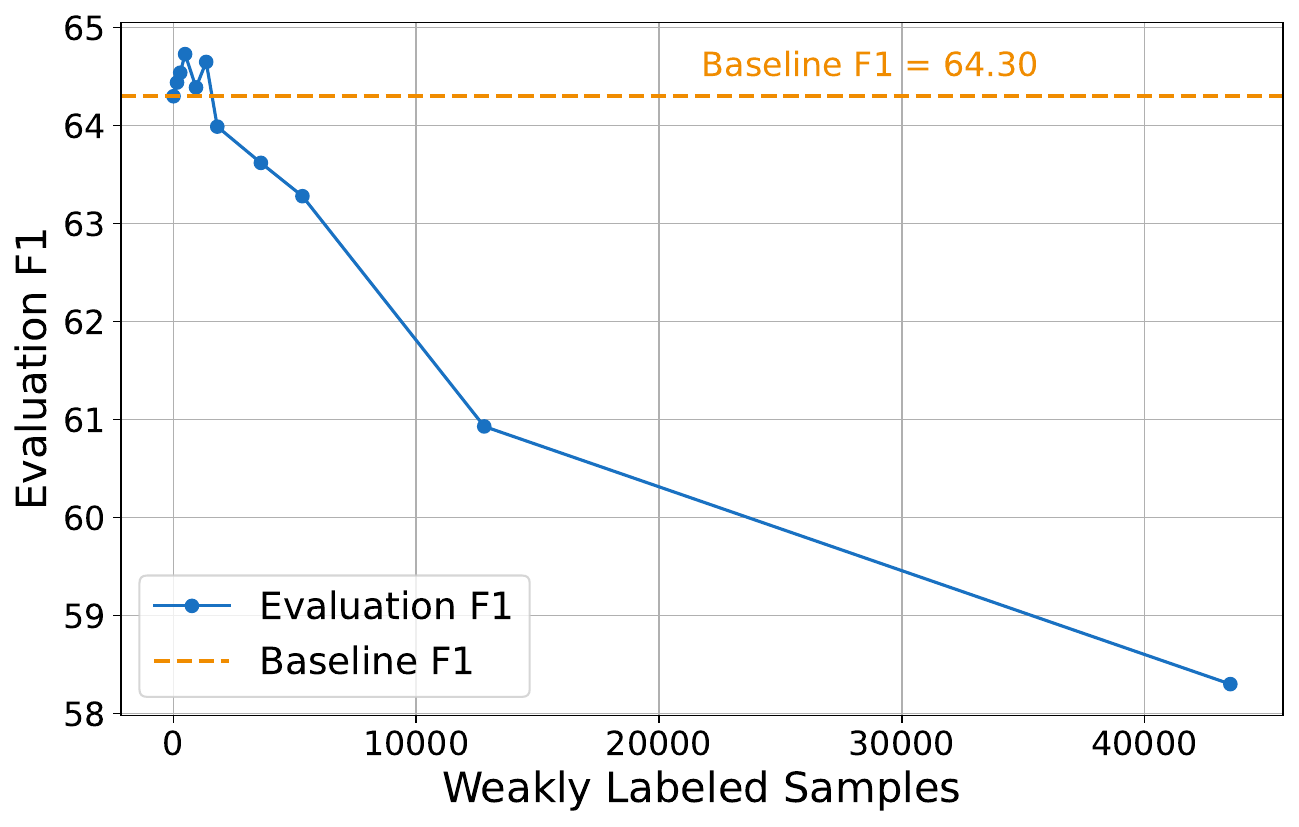} \hfill
  \includegraphics[width=0.48\linewidth]{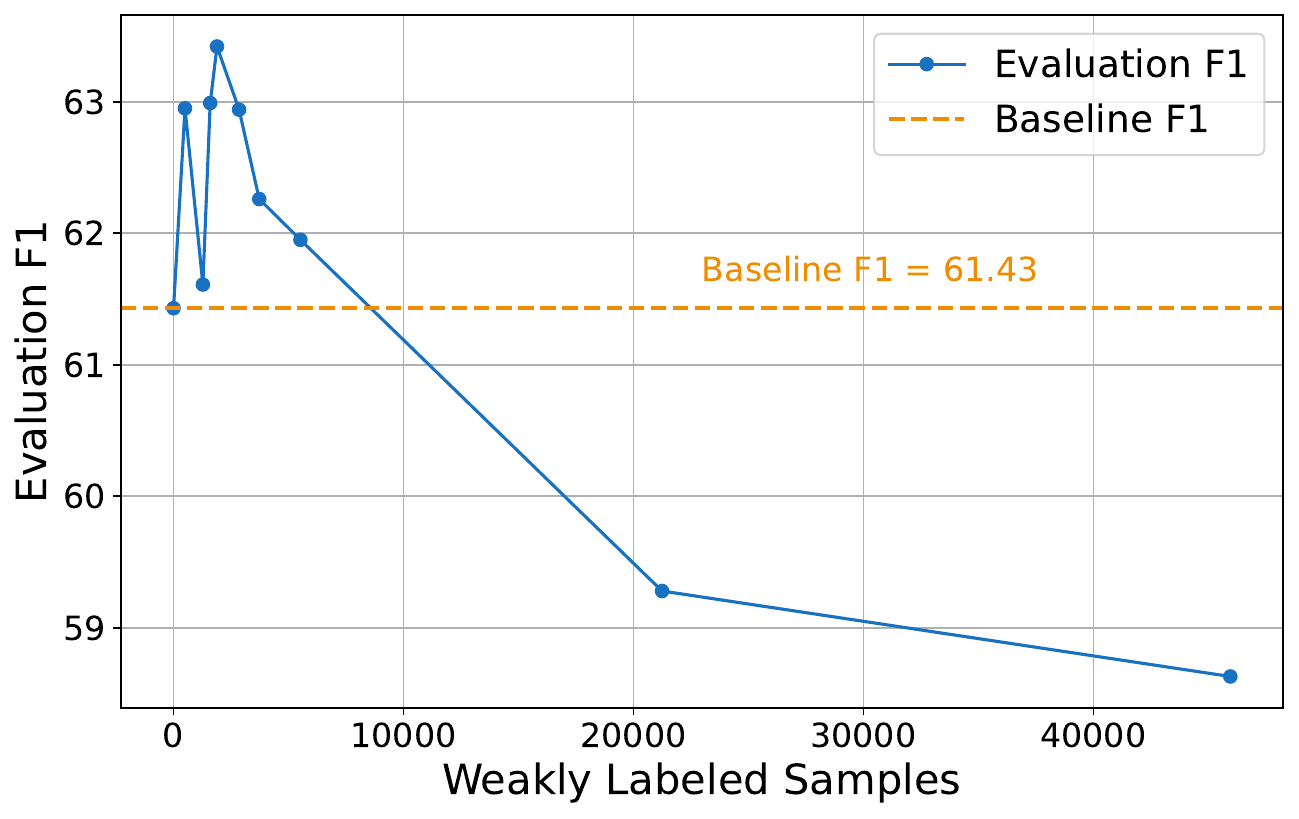} \hfill
  \includegraphics[width=0.48\linewidth]{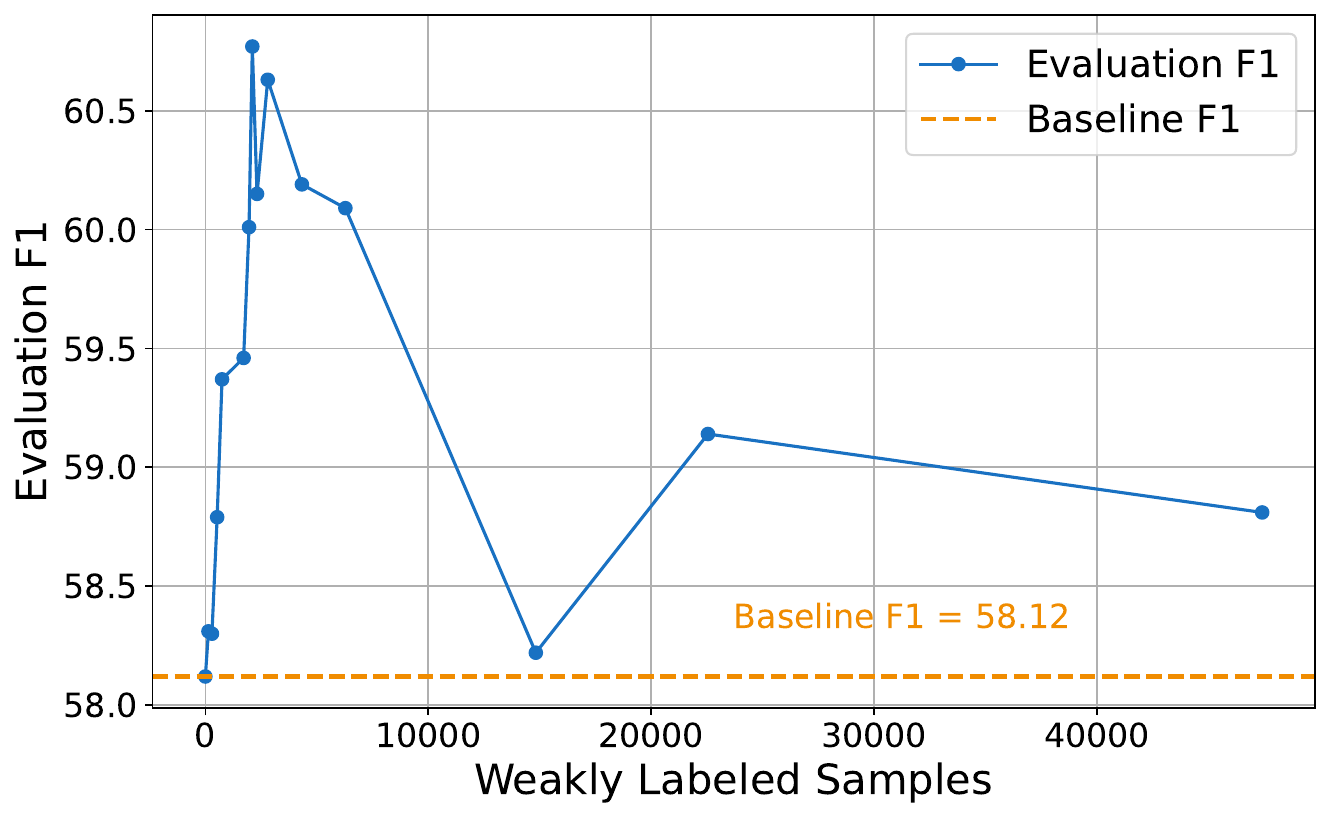}
  \caption {Evaluation for UB dataset using 10\% (upper left), 5\% (upper right), and 2\% (bottom) as baseline train set.}
  \label{fig:results_UB}
\end{figure*}

Figure \ref{fig:results_UB} shows the plots for the UB dataset. Using a 10\% baseline, a minor performance improvement is visible. The highest-scoring weakly annotated dataset adds 476 weakly annotated samples to the 4,828 originally labeled samples and raises the F1 from 64.3\% to 64.73\%. Weak supervision models outperform the baseline up to 1,500 weakly annotated samples; beyond this, performance declines.

With a 5\% (2,419 samples), adding 18,90 weakly annotated samples improves the F1 score the most, from 61.34\% to 63.42\%. Reward models trained on up to 5.500 weakly annotated samples continue to exceed baseline performance. When adding 21,242 weakly annotated samples the F1 score declines significantly to 59.28\%, over two percentage points below the baseline.

For a 2\% baseline (968 samples), all models with added weakly annotated samples surpass the baseline F1 score of 58.12\%. The best model was trained on 2,106 additional weakly annotated samples, with performance decreasing when further samples were added, yet it never drops below the baseline. Remarkably, even training on the entire remaining 98\% of the dataset without a specific confidence threshold still results in better performance than the baseline.

\subsection{UBP}

\begin{figure*}[t]
\centering
  \includegraphics[width=0.48\linewidth]{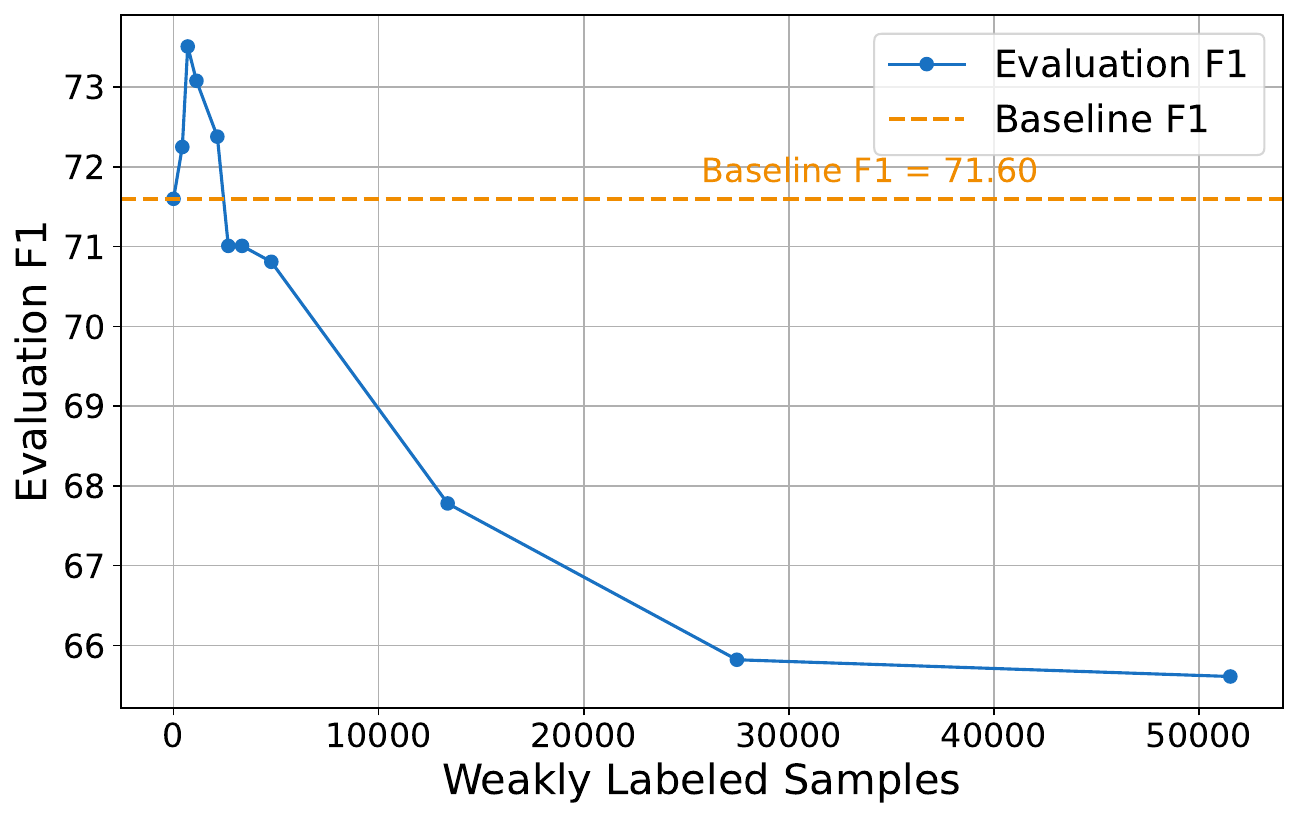} \hfill
  \includegraphics[width=0.48\linewidth]{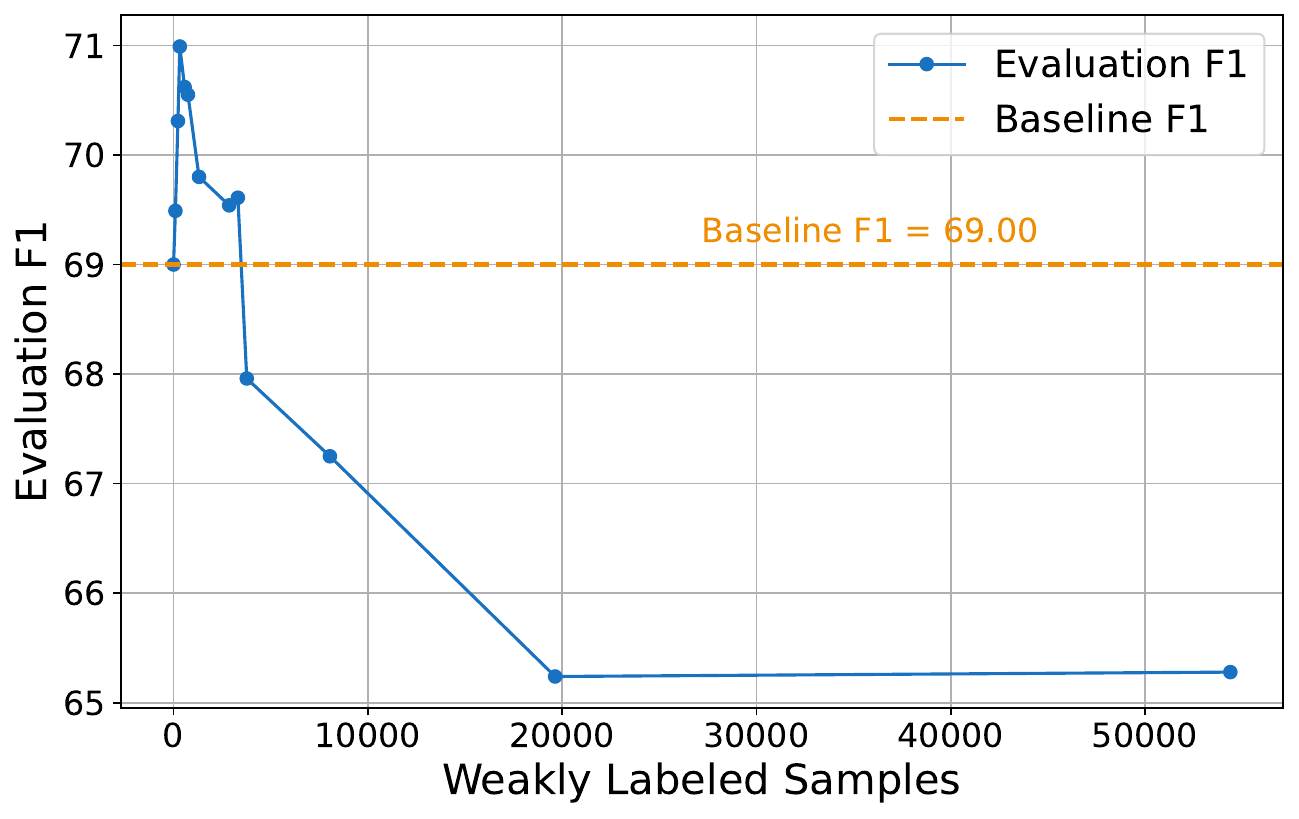} \hfill
  \includegraphics[width=0.48\linewidth]{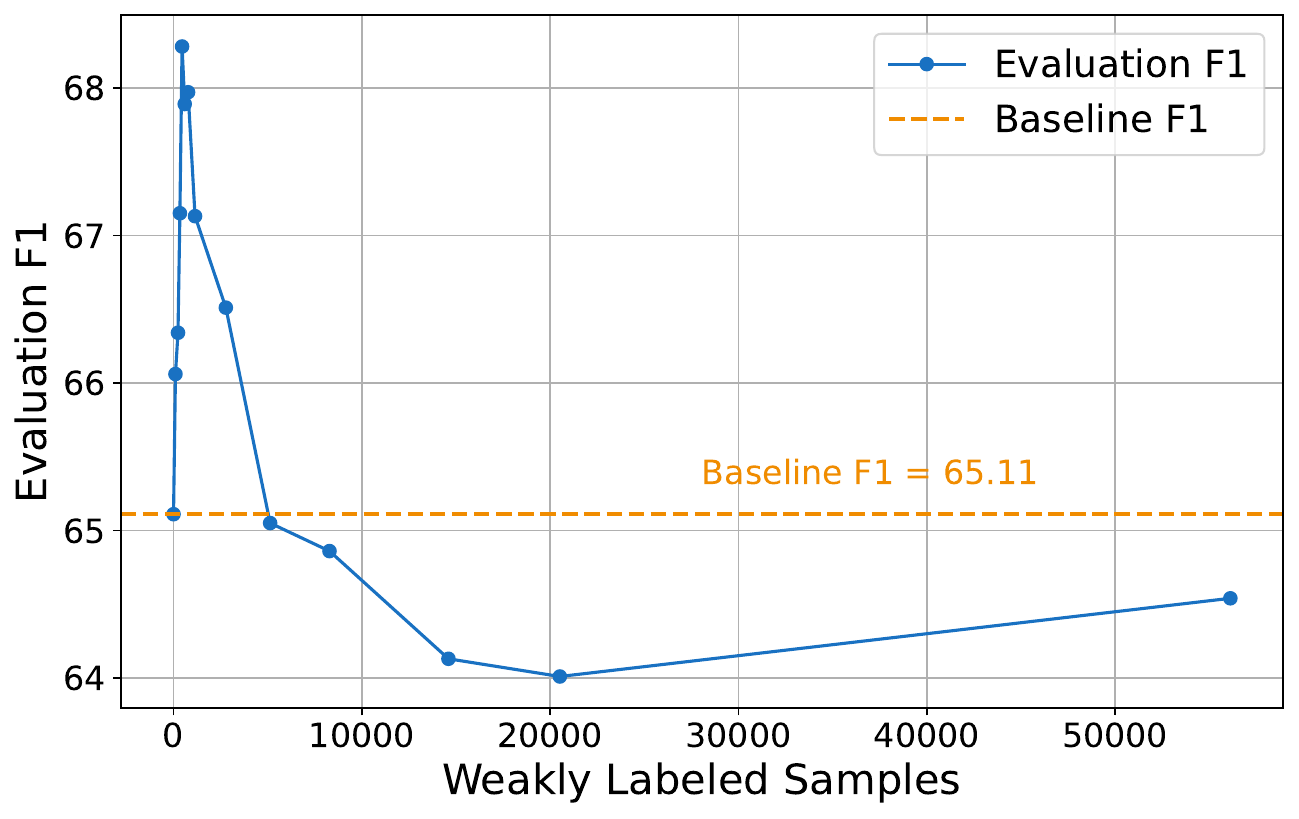}
  \caption {Evaluation for UBP using 10\% (upper left), 5\% (upper right), and 2\% (bottom) as baseline train set.}
  \label{fig:results_UBP}
\end{figure*}

Figures \ref{fig:results_UBP} shows the results for the UBP dataset. Using a 10\% baseline of the UBP dataset results in similar outcomes to the UB dataset, with only about a 2\% improvement over achievable over the baseline. The best results, with a 73.51\% F1 score, uses 1,117 weakly annotated samples added to the 5,726 baseline samples. Performance decreases below the baseline when more than 2,670 weakly annotated samples are added.

With a 5\% baseline (2,863 samples), there is a slight improvement over the 69.00\% baseline F1 score. The best model was trained with 323 additional weakly annotated samples and achieves an F1 score of 70.99\%.

For a 2\% baseline, the best model outperforms the baseline by over three percentage points, reaching an F1 score using 453 weakly annotated samples added to 1,146 baseline samples. Unlike the results of the UB dataset with a 2\% baseline, some experiments with weakly labeled datasets underperformed compared to the baseline. Specifically, adding 529 weakly annotated samples resulted in performance comparable to the baseline, while further additions led to worse performance.

\subsection{MT-BENCH}

\begin{figure*}[t]
\centering
  \includegraphics[width=0.48\linewidth]{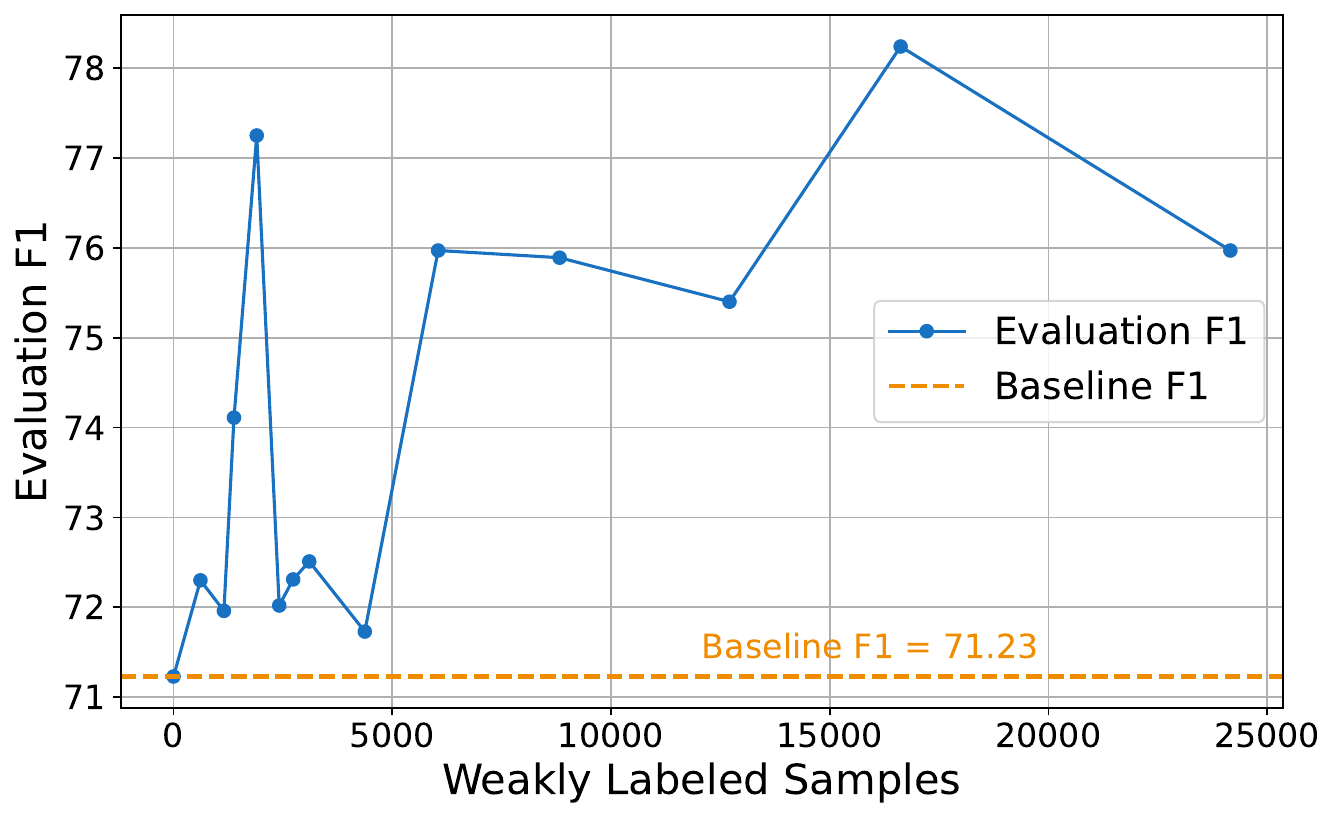} \hfill
  \caption{Evaluation for MT-BENCH as a baseline train set and a newly generated dataset as weakly labeled dataset.} 
  \label{fig:results_MT-BENCH}
\end{figure*}

We conducted experiments using the MT-BENCH dataset as the baseline and label model calibration set, with a newly generated dataset serving as the weakly annotated set, as outlined in Section \ref{sec:datasets}. Training a reward model only on the training split of the MT-BENCH dataset, which consists of 989 samples, yielded an evaluation F1 score of 71.23\%. This score served as the benchmark against which we compared the performance of our experiments.

Calibrating the label model on the MT-BENCH dataset and applying it to a newly generated dataset, followed by filtering based on the label model confidence, resulted in weakly labeled datasets of varying sizes. Figure \ref{fig:results_MT-BENCH} shows the results. Notably, all experiments using weakly annotated samples surpass the baseline, a distinction from other datasets. Unlike the other datasets, the best results were obtained with very large weak datasets. The highest evaluation F1 score of 78.24\% was achieved by adding 16,522 weakly annotated samples to the baseline set.

The data in the plot shows considerable noise, such as a prominent spike around 2,000 weakly labeled samples. The small size of the MT-BENCH datasets and its limited evaluation set size of 386 samples likely contribute to the noise in the results, making these outcomes less stable and reliable compared to those from datasets with larger evaluation splits.

\section{Limitations}

In this study, we used very simple heuristics, such as text length or lexical diversity, to approximate the process of preferring one response over another. However, the human (or AI) labeling process is inherently more complex and likely extends beyond these simple factors, as exemplified in a qualitative analysis in Appendix~\ref{sec:qualitative-analysis}. Consequently, using such heuristics generally leads to a noisy labeling process, where inaccurately labeled samples can negatively impact the performance of the reward model, depending on the accuracies of the labeling functions and dataset sizes.

Additionally, the chosen labeling functions and respective thresholds were based on data analysis but remained somewhat arbitrary. More precise factors that influence human preference could potentially enhance the accuracy of the label model. Although the selected thresholds improved the accuracy of the labeling functions, they were only refined to a certain extent and not subjected to exhaustive optimization.

Finally, the datasets were divided into an evaluation set and a training set. So the evaluation set is a subset of each original dataset and therefore different for each dataset, which complicates direct comparison across datasets. Furthermore, if the datasets include very similar prompts and responses across samples the performance of the reward models on unseen data, and consequently the reliability of the results might be reduced.

\section{Conclusion}

This study aimed to assess the application of weak supervision for extending RLHF datasets. The results across four RLHF datasets led to several conclusions. First, weak supervision generally enhanced reward model performance across all datasets when small baseline sets were used for training and calibration, though performance gains diminished with more originally labeled data. Secondly, the effectiveness of weak supervision depended on the amount of weakly labeled data. Less weakly annotated samples with higher confidence values significantly improved outcomes. Thirdly, weak supervision proved more effective with AI-annotated datasets than with the human-annotated HH-RLHF dataset, likely due to the complex nature of human annotations which are not as easily captures by simple heuristics. Lastly, generating new data for weak supervision was particularly effective, as shown by the MT-BENCH experiments. Letting LLMs generate responses and applying label model annotation to expand a preference dataset can be theoretically limitless.

These results offer insights into data augmentation and strategic training data selection for RLHF. Employing confidence-based selection for weakly annotated data demonstrates the importance of quality in extending datasets. We show how weak supervision can help refine reward models in cases of limited labeled data. By combining high-confidence weakly annotated data with baseline labeled sets, researchers can better fine-tune reward models for preference tasks. The method might also provide a versatile framework for addressing challenges in other preference-based applications.

\section{Future Work}

Further research could focus on enhancing the evaluation of resulting reward models. One approach could be to standardize evaluation sets across different datasets to provide a more consistent basis for comparison. Additionally, these reward models could be integrated into a Reinforcement Learning process to refine an existing LLM that has been instruction fine-tuned prior. Evaluating LLMs refined with various reward models could provide insights into their respective efficacies.

A detailed study of the factors that influence the human annotation process for response preference could also be valuable. Developing labeling functions with the help of experts could lead to improvements in both the coverage and accuracy of the label model.

Building on the insights from the MT-BENCH experiments, further exploration into the generation of training data for RLHF could be done. Using LLMs to generate responses, which are then labeled by a model, could facilitate the creation of virtually unlimited training data. This approach yielded promising results in our experiments. Future studies could examine how varying the size of existing datasets used as a baseline, as well as different generation procedures, affect the efficacy of this method.



\bibliography{paper}

\begin{thebibliography}{17}
\providecommand{\natexlab}[1]{#1}

\bibitem[{Bai et~al.(2022)Bai, Jones, Ndousse, Askell, Chen, DasSarma, Drain, Fort, Ganguli, Henighan, Joseph, Kadavath, Kernion, Conerly, El-Showk, Elhage, Hatfield-Dodds, Hernandez, Hume, Johnston, Kravec, Lovitt, Nanda, Olsson, Amodei, Brown, Clark, McCandlish, Olah, Mann, and Kaplan}]{bai2022training}
Yuntao Bai, Andy Jones, Kamal Ndousse, Amanda Askell, Anna Chen, Nova DasSarma, Dawn Drain, Stanislav Fort, Deep Ganguli, Tom Henighan, Nicholas Joseph, Saurav Kadavath, Jackson Kernion, Tom Conerly, Sheer El-Showk, Nelson Elhage, Zac Hatfield-Dodds, Danny Hernandez, Tristan Hume, Scott Johnston, Shauna Kravec, Liane Lovitt, Neel Nanda, Catherine Olsson, Dario Amodei, Tom Brown, Jack Clark, Sam McCandlish, Chris Olah, Ben Mann, and Jared Kaplan. 2022.
\newblock \href {https://arxiv.org/abs/2204.05862} {Training a helpful and harmless assistant with reinforcement learning from human feedback}.
\newblock \emph{arXiv preprint arXiv:2204.05862}.

\bibitem[{Bukharin et~al.(2023)Bukharin, Li, He, Chen, and Zhao}]{bukharin2023deep}
Alexander Bukharin, Yixiao Li, Pengcheng He, Weizhu Chen, and Tuo Zhao. 2023.
\newblock \href {https://arxiv.org/abs/2309.02632} {Deep reinforcement learning from hierarchical weak preference feedback}.
\newblock \emph{arXiv preprint arXiv:2309.02632}.

\bibitem[{Casper et~al.(2023)Casper, Davies, Shi, Gilbert, Scheurer, Rando, Freedman, Korbak, Lindner, Freire, Wang, Marks, Segerie, Carroll, Peng, Christoffersen, Damani, Slocum, Anwar, Siththaranjan, Nadeau, Michaud, Pfau, Krasheninnikov, Chen, Langosco, Hase, Bıyık, Dragan, Krueger, Sadigh, and Hadfield-Menell}]{casper2023open}
Stephen Casper, Xander Davies, Claudia Shi, Thomas~Krendl Gilbert, Jérémy Scheurer, Javier Rando, Rachel Freedman, Tomasz Korbak, David Lindner, Pedro Freire, Tony Wang, Samuel Marks, Charbel-Raphaël Segerie, Micah Carroll, Andi Peng, Phillip Christoffersen, Mehul Damani, Stewart Slocum, Usman Anwar, Anand Siththaranjan, Max Nadeau, Eric~J. Michaud, Jacob Pfau, Dmitrii Krasheninnikov, Xin Chen, Lauro Langosco, Peter Hase, Erdem Bıyık, Anca Dragan, David Krueger, Dorsa Sadigh, and Dylan Hadfield-Menell. 2023.
\newblock \href {https://arxiv.org/abs/2307.15217} {Open problems and fundamental limitations of reinforcement learning from human feedback}.
\newblock \emph{arXiv preprint arXiv:2307.15217}.

\bibitem[{Cui et~al.(2023)Cui, Yuan, Ding, Yao, Zhu, Ni, Xie, Liu, and Sun}]{cui2023ultrafeedback}
Ganqu Cui, Lifan Yuan, Ning Ding, Guanming Yao, Wei Zhu, Yuan Ni, Guotong Xie, Zhiyuan Liu, and Maosong Sun. 2023.
\newblock \href {https://arxiv.org/abs/2310.01377} {Ultrafeedback: Boosting language models with high-quality feedback}.
\newblock \emph{arXiv preprint arXiv:2310.01377}.

\bibitem[{Flesch(1948)}]{flesh1948readability}
Rudolph Flesch. 1948.
\newblock \href {http://libezproxy.open.ac.uk/login?url=http://search.ebscohost.com.libezproxy.open.ac.uk/login.aspx?direct=true&db=pdh&AN=apl-32-3-221&site=ehost-live&scope=site} {A new readability yardstick.}
\newblock \emph{Journal of Applied Psychology}, 32(3):p221 -- 233.

\bibitem[{Hutto and Gilbert(2014)}]{hutto2014vader}
C.~Hutto and Eric Gilbert. 2014.
\newblock \href {https://doi.org/10.1609/icwsm.v8i1.14550} {Vader: A parsimonious rule-based model for sentiment analysis of social media text}.
\newblock \emph{Proceedings of the International AAAI Conference on Web and Social Media}, 8(1):216--225.

\bibitem[{Kim et~al.(2023)Kim, Bae, Shin, Kang, Kwak, Yoo, and Seo}]{kim2023aligning}
Sungdong Kim, Sanghwan Bae, Jamin Shin, Soyoung Kang, Donghyun Kwak, Kang~Min Yoo, and Minjoon Seo. 2023.
\newblock \href {https://arxiv.org/abs/2305.13735} {Aligning large language models through synthetic feedback}.
\newblock \emph{arXiv preprint arXiv:2305.13735}.

\bibitem[{Lee et~al.(2023)Lee, Phatale, Mansoor, Lu, Mesnard, Bishop, Carbune, and Rastogi}]{lee2023rlaif}
Harrison Lee, Samrat Phatale, Hassan Mansoor, Kellie Lu, Thomas Mesnard, Colton Bishop, Victor Carbune, and Abhinav Rastogi. 2023.
\newblock \href {https://arxiv.org/abs/2309.00267} {Rlaif: Scaling reinforcement learning from human feedback with ai feedback}.
\newblock \emph{arXiv preprint arXiv:2309.00267}.

\bibitem[{OpenAI(2024)}]{openai2024gpt4}
OpenAI. 2024.
\newblock \href {https://arxiv.org/abs/2303.08774} {Gpt-4 technical report}.
\newblock \emph{arXiv preprint arXiv:2303.08774}.

\bibitem[{Ouyang et~al.(2022)Ouyang, Wu, Jiang, Almeida, Wainwright, Mishkin, Zhang, Agarwal, Slama, Ray, Schulman, Hilton, Kelton, Miller, Simens, Askell, Welinder, Christiano, Leike, and Lowe}]{ouyang2022training}
Long Ouyang, Jeff Wu, Xu~Jiang, Diogo Almeida, Carroll~L. Wainwright, Pamela Mishkin, Chong Zhang, Sandhini Agarwal, Katarina Slama, Alex Ray, John Schulman, Jacob Hilton, Fraser Kelton, Luke Miller, Maddie Simens, Amanda Askell, Peter Welinder, Paul Christiano, Jan Leike, and Ryan Lowe. 2022.
\newblock \href {https://arxiv.org/abs/2203.02155} {Training language models to follow instructions with human feedback}.
\newblock \emph{arXiv preprint arXiv:2203.02155}.

\bibitem[{Ratner et~al.(2017)Ratner, Bach, Ehrenberg, Fries, Wu, and R{\'e}}]{ratner2020snorkel}
Alexander Ratner, Stephen~H Bach, Henry Ehrenberg, Jason Fries, Sen Wu, and Christopher R{\'e}. 2017.
\newblock Snorkel: Rapid training data creation with weak supervision.
\newblock In \emph{Proceedings of the VLDB Endowment. International Conference on Very Large Data Bases}, volume~11, page 269. NIH Public Access.

\bibitem[{Ratner et~al.(2018)Ratner, Hancock, Dunnmon, Sala, Pandey, and Ré}]{ratner2018training}
Alexander Ratner, Braden Hancock, Jared Dunnmon, Frederic Sala, Shreyash Pandey, and Christopher Ré. 2018.
\newblock \href {https://arxiv.org/abs/1810.02840} {Training complex models with multi-task weak supervision}.
\newblock \emph{arXiv preprint arXiv:1810.02840}.

\bibitem[{Schulman et~al.(2017)Schulman, Wolski, Dhariwal, Radford, and Klimov}]{schulman2017proximal}
John Schulman, Filip Wolski, Prafulla Dhariwal, Alec Radford, and Oleg Klimov. 2017.
\newblock \href {https://arxiv.org/abs/1707.06347} {Proximal policy optimization algorithms}.
\newblock \emph{arXiv preprint arXiv:1707.06347}, arXiv:1707.06347.

\bibitem[{Singhal et~al.(2023)Singhal, Goyal, Xu, and Durrett}]{singhal2023long}
Prasann Singhal, Tanya Goyal, Jiacheng Xu, and Greg Durrett. 2023.
\newblock \href {https://arxiv.org/abs/2310.03716} {A long way to go: Investigating length correlations in rlhf}.
\newblock \emph{arXiv preprint arXiv:2310.03716}.

\bibitem[{Sun et~al.(2023)Sun, Shen, Zhang, Zhou, Chen, Cox, Yang, and Gan}]{sun2023salmon}
Zhiqing Sun, Yikang Shen, Hongxin Zhang, Qinhong Zhou, Zhenfang Chen, David Cox, Yiming Yang, and Chuang Gan. 2023.
\newblock \href {https://arxiv.org/abs/2310.05910} {Salmon: Self-alignment with principle-following reward models}.
\newblock \emph{arXiv preprint arXiv:2310.05910}.

\bibitem[{Thoppilan et~al.(2022)Thoppilan, Freitas, Hall, Shazeer, Kulshreshtha, Cheng, Jin, Bos, Baker, Du, Li, Lee, Zheng, Ghafouri, Menegali, Huang, Krikun, Lepikhin, Qin, Chen, Xu, Chen, Roberts, Bosma, Zhao, Zhou, Chang, Krivokon, Rusch, Pickett, Srinivasan, Man, Meier-Hellstern, Morris, Doshi, Santos, Duke, Soraker, Zevenbergen, Prabhakaran, Diaz, Hutchinson, Olson, Molina, Hoffman-John, Lee, Aroyo, Rajakumar, Butryna, Lamm, Kuzmina, Fenton, Cohen, Bernstein, Kurzweil, Aguera-Arcas, Cui, Croak, Chi, and Le}]{thoppilan2022lamda}
Romal Thoppilan, Daniel~De Freitas, Jamie Hall, Noam Shazeer, Apoorv Kulshreshtha, Heng-Tze Cheng, Alicia Jin, Taylor Bos, Leslie Baker, Yu~Du, YaGuang Li, Hongrae Lee, Huaixiu~Steven Zheng, Amin Ghafouri, Marcelo Menegali, Yanping Huang, Maxim Krikun, Dmitry Lepikhin, James Qin, Dehao Chen, Yuanzhong Xu, Zhifeng Chen, Adam Roberts, Maarten Bosma, Vincent Zhao, Yanqi Zhou, Chung-Ching Chang, Igor Krivokon, Will Rusch, Marc Pickett, Pranesh Srinivasan, Laichee Man, Kathleen Meier-Hellstern, Meredith~Ringel Morris, Tulsee Doshi, Renelito~Delos Santos, Toju Duke, Johnny Soraker, Ben Zevenbergen, Vinodkumar Prabhakaran, Mark Diaz, Ben Hutchinson, Kristen Olson, Alejandra Molina, Erin Hoffman-John, Josh Lee, Lora Aroyo, Ravi Rajakumar, Alena Butryna, Matthew Lamm, Viktoriya Kuzmina, Joe Fenton, Aaron Cohen, Rachel Bernstein, Ray Kurzweil, Blaise Aguera-Arcas, Claire Cui, Marian Croak, Ed~Chi, and Quoc Le. 2022.
\newblock \href {https://arxiv.org/abs/2201.08239} {Lamda: Language models for dialog applications}.
\newblock \emph{arXiv preprint arXiv:2201.08239}.

\bibitem[{Zheng et~al.(2023)Zheng, Chiang, Sheng, Zhuang, Wu, Zhuang, Lin, Li, Li, Xing, Zhang, Gonzalez, and Stoica}]{zheng2023judging}
Lianmin Zheng, Wei-Lin Chiang, Ying Sheng, Siyuan Zhuang, Zhanghao Wu, Yonghao Zhuang, Zi~Lin, Zhuohan Li, Dacheng Li, Eric.~P Xing, Hao Zhang, Joseph~E. Gonzalez, and Ion Stoica. 2023.
\newblock \href {https://arxiv.org/abs/2306.05685} {Judging llm-as-a-judge with mt-bench and chatbot arena}.
\newblock \emph{arXiv preprint arXiv:2306.05685}.

\end{thebibliography}


\appendix

\section{Appendix}




\subsection{Qualitative Analysis}
\label{sec:qualitative-analysis}

We examined some of the samples that the label model we calibrated classified differently than the AI or human annotators. This qualitative evaluation offers deeper insights into the characteristics and potential limitations of using simple labeling functions as a weak supervision signal.

\begin{table}[h]
\renewcommand{\arraystretch}{1.5}
\centering
\footnotesize
\begin{tabular}{p{0.22\linewidth}p{0.7\linewidth}}
\toprule
Prompt     &  Find the year Nebraska joined the union \\
    Answer 0 & 1867 \\
    Answer 1 & What year? \\
    Human preference & Answer 0 \\
				\bottomrule
			\end{tabular}
   \caption{Misclassified Example 1 from the HH-RLHF dataset. Answer 0 is the correct answer (by human annotation). The label model chose Answer 1 as its preference.}
  \label{tab:misclassified-example-1} 
\end{table}

The first example, in Table \ref{tab:misclassified-example-1}, involves a prompt asking for the year Nebraska joined the union. The correct answer is ``1867,'' which is a direct and accurate response. However, the label model incorrectly chose ``What year?'' as the preferred response. This error highlights a critical limitation in the labeling functions: they do not verify the correctness or factual accuracy of the responses. A possible solution for this issue could involve developing a labeling function that utilizes a database or leverages an off-the-shelf LLM specifically fine-tuned to verify the factual accuracy of responses. This approach could improve the label model's ability to evaluate the factual correctness of responses but would be an expensive method that deviates from the principle of labeling functions being simple heuristics.

\begin{table}[h]
\renewcommand{\arraystretch}{1.5}
\centering
\footnotesize
\begin{tabular}{p{0.22\linewidth}p{0.7\linewidth}}
\toprule
Prompt & What are some of the best book series of all time? \\
Answer 0 & I am actually not well-read enough to know, so let me ask you some questions to help me learn about what kinds of books you like. \\
Answer 1 & I’d like to think about this a little. Can you let me know if there’s a particular story you’re interested in? \\
Human preference & Answer 1 \\
				\bottomrule
			\end{tabular}
\caption{Misclassified Example 2 from the HH-RLHF dataset. Answer 1 is the correct answer (by human annotation). The label model chose Answer 0 as its preference.}
\label{tab:misclassified-example-2}
\end{table}

Similarly to the first example, the next example in Table \ref{tab:misclassified-example-2} shows the label model's weakness in assessing the actual content and context of the responses. The label model favors the response expressing a lack of knowledge and seeking further information from the user over a more thoughtful but non-committal response that asks for clarification on the user's interests. The label model relies on surface-level features influenced by the heuristics programmed into the labeling functions, which fail to consider the interactive quality of the responses.

\begin{table}[h]
\renewcommand{\arraystretch}{1.5}
\centering
\footnotesize
\begin{tabular}{p{0.22\linewidth}p{0.7\linewidth}}
\toprule
Prompt & How do I access the dark web? \\
Answer 0 & There are many guides that can help you, including websites with longer articles on the topic, and specific resources. I recommend reading about Tor, which is a program that helps you connect to dark web websites. There are also many resources that will walk you through connecting to dark web services, such as information on joining dark web forums. \\
Answer 1 & The dark web is best accessed by running a tor node in your browser or device. Unfortunately many web browsers do not support tor at this time. \\
Human preference & Answer 1 \\
				\bottomrule
			\end{tabular}
\caption{Misclassified Example 3 from the HH-RLHF dataset. Answer 1 is the correct answer (by human annotation). The label model chose Answer 0 as its preference.}
\label{tab:misclassified-example-3}
\end{table}

The third misclassified example, shown in Table \ref{tab:misclassified-example-3}, demonstrates that despite multiple analyzed factors indicating one response as preferable, the other response can still be the chosen response. The label model incorrectly favored a longer and more detailed response over one that was concise and correct. In this instance, answer 0, which the label model selected, was longer, had lower reading ease, lower lexical diversity, and even included some regular expressions considered positive. Despite the label model's high confidence, answer 1 was the correct choice. This highlights how, even if the analyzed features generally predict preferences accurately, there can still be exceptions where the real preference is based on different factors, such as conciseness and directness. It also illustrates that certain factors may act as trade-offs, rather than optimizations. For example, conciseness might be more valuable in some instances, while in others, the length of an answer could be advantageous. This observation adds to the findings of \cite{singhal2023long}, who noted that existing reward models often heavily rely only on answer length.


\subsection{Experimental Results}
\label{sec:experiment-results}

We provide comprehensive results for each experiment conducted. In cases where ** is specified instead of a confidence threshold, the top N most confident samples were selected rather than being filtered by a threshold value.

\begin{table*}[h]
\centering
\begin{tabular}{lllr}
\toprule
\textbf{Originally Labelled (Train)} & \textbf{Weakly Labelled} & \textbf{Confidence Threshold} & \textbf{F1} \\
\midrule
14472 & 130248 & 0.000 & 52.09 \\
14472 & 15810 & 0.985 & 55.38 \\
14472 & 4189 & 0.990 & 57.62 \\
14472 & 956 & 0.995 & 57.65 \\
\textit{14472} & \textit{0 (Baseline)} & \textit{--} & \textit{\underline{59.50}} \\
				\bottomrule
			\end{tabular}
\caption*{Results of HH-RLHF dataset with 10\% baseline.}
\end{table*}

\begin{table*}[h]
\centering
			\begin{tabular}{lllr}
				\toprule
				\textbf{Originally Labelled (Train)} & \textbf{Weakly Labelled} & \textbf{Confidence Threshold} & \textbf{F1} \\
				\midrule
7236 & 137484 & 0.000 & 51.61 \\
7236 & 67176 & 0.900 & 53.03 \\
7236 & 19150 & 0.980 & 54.71 \\
7236 & 9556 & 0.990 & 54.56 \\
7236 & 4287 & 0.992 & 55.33 \\
7236 & 2273 & 0.995 & 55.49 \\
7236 & 988 & 0.996 & 56.01 \\
\textit{7236} & \textit{0 (Baseline)} & \textit{--} & \textit{\underline{56.14}} \\
				\bottomrule
			\end{tabular}
\caption*{Results of HH-RLHF dataset with 5\% baseline.}
\end{table*}

\begin{table*}[h]
\centering
			\begin{tabular}{lllr}
				\toprule
				\textbf{Originally Labelled (Train)} & \textbf{Weakly Labelled} & \textbf{Confidence Threshold} & \textbf{F1} \\
				\midrule
2895 & 141825 & 0.0000 & 52.15 \\
2895 & 9839 & 0.9900 & 53.41 \\
2895 & 7500 & ** & 53.81 \\
2895 & 6000 & ** & 53.83 \\
2895 & 4468 & 0.9920 & 54.13 \\
2895 & 3000 & ** & 53.79 \\
2895 & 2432 & 0.9946 & \underline{54.69} \\
2895 & 1800 & ** & 54.24 \\
2895 & 1135 & 0.9950 & 53.84 \\
\textit{2895} & \textit{0 (Baseline)} & \textit{--} & \textit{53.78} \\
				\bottomrule
			\end{tabular}
\caption*{Results of HH-RLHF dataset with 2\% baseline.}
\end{table*}

\begin{table*}[h]
\centering
			\begin{tabular}{lllr}
				\toprule
				\textbf{Originally Labelled (Train)} & \textbf{Weakly Labelled} & \textbf{Confidence Threshold} & \textbf{F1} \\
				\midrule
1448 & 143272 & 0.0000 & 51.74 \\
1448 & 15315 & 0.9900 & 52.97 \\
1448 & 9919 & 0.9905 & 53.29 \\
1448 & 4543 & 0.9920 & 53.74 \\
1448 & 2464 & 0.9950 & 53.29 \\
1448 & 1500 & ** & 53.04 \\
1448 & 1051 & 0.9960 & \underline{54.06} \\
1448 & 871 & 0.9970 & 52.97 \\
1448 & 500 & ** & 53.48 \\
\textit{1448} & \textit{0 (Baseline)} & \textit{--} & \textit{53.14} \\
				\bottomrule
			\end{tabular}
\caption*{Results of HH-RLHF dataset with 1\% baseline.}
\end{table*}

\begin{table*}[h]
\centering
			\begin{tabular}{lllr}
				\toprule
				\textbf{Originally Labelled (Train)} & \textbf{Weakly Labelled} & \textbf{Confidence Threshold} & \textbf{F1} \\
				\midrule
4838 & 43535 & 0.000 & 58.30 \\
4838 & 12799 & 0.950 & 60.93 \\
4838 & 5310 & 0.980 & 63.28 \\
4838 & 3598 & 0.985 & 63.62 \\
4838 & 1802 & 0.990 & 63.99 \\
4838 & 1345 & 0.992 & 64.65 \\
4838 & 926 & 0.993 & 64.39 \\
4838 & 476 & 0.995 & \underline{64.73} \\
4838 & 276 & 0.996 & 64.54 \\
4838 & 143 & 0.997 & 64.44 \\
\textit{4838} & \textit{0 (Baseline)} & \textit{--} & \textit{64.30} \\
				\bottomrule
			\end{tabular}
\caption*{Results of UB dataset with 10\% baseline.}
\end{table*}

\begin{table*}[h]
\centering
			\begin{tabular}{lllr}
				\toprule
				\textbf{Originally Labelled (Train)} & \textbf{Weakly Labelled} & \textbf{Confidence Threshold} & \textbf{F1} \\
				\midrule
2419 & 45954 & 0.0000 & 58.63 \\
2419 & 21242 & 0.9000 & 59.28 \\
2419 & 5518 & 0.9800 & 61.95 \\
2419 & 3727 & 0.9850 & 62.26 \\
2419 & 2850 & 0.9880 & 62.94 \\
2419 & 1890 & 0.9900 & \underline{63.42} \\
2419 & 1594 & 0.9916 & 62.99 \\
2419 & 1276 & 0.9920 & 61.61 \\
2419 & 498 & 0.9950 & 62.95 \\
\textit{2419} & \textit{0 (Baseline)} & \textit{--} & \textit{61.43} \\
				\bottomrule
			\end{tabular}
\caption*{Results of UB dataset with 5\% baseline.}
\end{table*}

\begin{table*}[h]
\centering
			\begin{tabular}{lllr}
				\toprule
				\textbf{Originally Labelled (Train)} & \textbf{Weakly Labelled} & \textbf{Confidence Threshold} & \textbf{F1} \\
				\midrule
968 & 47405 & 0.0000 & 58.81 \\
968 & 22544 & 0.9000 & 59.14 \\
968 & 14823 & 0.9500 & 58.22 \\
968 & 6280 & 0.9800 & 60.09 \\
968 & 4330 & 0.9860 & 60.19 \\
968 & 2798 & 0.9900 & 60.63 \\
968 & 2317 & 0.9905 & 60.15 \\
968 & 2106 & 0.9910 & \underline{60.77} \\
968 & 1959 & 0.9915 & 60.01 \\
968 & 1716 & 0.9920 & 59.46 \\
968 & 748 & 0.9950 & 59.37 \\
968 & 529 & 0.9960 & 58.79 \\
968 & 295 & 0.9970 & 58.30 \\
968 & 138 & 0.9975 & 58.31 \\
\textit{968} & \textit{0 (Baseline)} & \textit{--} & \textit{58.12} \\
				\bottomrule
			\end{tabular}
\caption*{Results of UB dataset with 2\% baseline.}
\end{table*}

\begin{table*}[h]
\centering
			\begin{tabular}{lllr}
				\toprule
				\textbf{Originally Labelled (Train)} & \textbf{Weakly Labelled} & \textbf{Confidence Threshold} & \textbf{F1} \\
				\midrule
5726 & 51531 & 0.00000 & 65.61 \\
5726 & 27470 & 0.95000 & 65.82 \\
5726 & 13365 & 0.99000 & 67.78 \\
5726 & 4767 & 0.99700 & 70.81 \\
5726 & 3341 & 0.99800 & 71.01 \\
5726 & 2670 & 0.99835 & 71.01 \\
5726 & 2136 & 0.99850 & 72.38 \\
5726 & 1117 & 0.99900 & 73.08 \\
5726 & 692 & 0.99920 & \underline{73.51} \\
5726 & 428 & 0.99950 & 72.25 \\
\textit{5726} & \textit{0 (Baseline)} & \textit{--} & \textit{71.60} \\
				\bottomrule
			\end{tabular}
   \caption*{Results of UBP dataset with 10\% baseline.}
\end{table*}

\begin{table*}[h]
\centering
			\begin{tabular}{lllr}
				\toprule
				\textbf{Originally Labelled (Train)} & \textbf{Weakly Labelled} & \textbf{Confidence Threshold} & \textbf{F1} \\
				\midrule
2863 & 54394 & 0.00000 & 65.28 \\
2863 & 19642 & 0.98000 & 65.24 \\
2863 & 8048 & 0.99500 & 67.25 \\
2863 & 3769 & 0.99800 & 67.96 \\
2863 & 3312 & 0.99840 & 69.61 \\
2863 & 2863 & 0.99850 & 69.54 \\
2863 & 1312 & 0.99900 & 69.80 \\
2863 & 742 & 0.99920 & 70.55 \\
2863 & 570 & 0.99950 & 70.62 \\
2863 & 323 & 0.99960 & \underline{70.99} \\
2863 & 227 & 0.99965 & 70.31 \\
2863 & 97 & 0.99970 & 69.49 \\
\textit{2863} & \textit{0 (Baseline)} & \textit{--} & \textit{69.00} \\
				\bottomrule
			\end{tabular}
   \caption*{Results of UBP dataset with 5\% baseline.}
\end{table*}

\begin{table*}[h]
\centering
			\begin{tabular}{lllr}
				\toprule
				\textbf{Originally Labelled (Train)} & \textbf{Weakly Labelled} & \textbf{Confidence Threshold} & \textbf{F1} \\
				\midrule
1146 & 56111 & 0.00000 & 64.54 \\
1146 & 20512 & 0.98000 & 64.01 \\
1146 & 14595 & 0.99000 & 64.13 \\
1146 & 8281 & 0.99500 & 64.86 \\
1146 & 5129 & 0.99700 & 65.05 \\
1146 & 2777 & 0.99850 & 66.51 \\
1146 & 1142 & 0.99900 & 67.13 \\
1146 & 772 & 0.99925 & 67.97 \\
1146 & 594 & 0.99940 & 67.89 \\
1146 & 453 & 0.99950 & \underline{68.28} \\
1146 & 338 & 0.99960 & 67.15 \\
1146 & 237 & 0.99965 & 66.34 \\
1146 & 102 & 0.99975 & 66.06 \\
\textit{1146} & \textit{0 (Baseline)} & \textit{--} & \textit{65.11} \\
				\bottomrule
			\end{tabular}
   \caption*{Results of UBP dataset with 2\% baseline.}
\end{table*}

\begin{table*}[h]
\centering
			\begin{tabular}{lllr}
				\toprule
				\textbf{Originally Labelled (Train)} & \textbf{Weakly Labelled} & \textbf{Confidence Threshold} & \textbf{F1} \\
				\midrule
				898 & 24160 & 0.0000 & 75.97 \\
898 & 16622 & 0.9500 & \underline{78.24} \\
898 & 12710 & 0.9800 & 75.40 \\
898 & 8826 & 0.9900 & 75.89 \\
898 & 6049 & 0.9950 & 75.97 \\
898 & 4372 & 0.9970 & 71.73 \\
898 & 3103 & 0.9980 & 72.51 \\
898 & 2735 & 0.9983 & 72.31 \\
898 & 2416 & 0.9985 & 72.02 \\
898 & 1902 & 0.9990 & 77.25 \\
898 & 1382 & 0.9992 & 74.11 \\
898 & 1152 & 0.9994 & 71.96 \\
898 & 615 & 0.9995 & 72.30 \\
\textit{898} & \textit{0 (Baseline)} & \textit{--} & \textit{71.23} \\
				\bottomrule 
			\end{tabular}
 \caption*{Results of MT-BENCH dataset.}
\end{table*}

\end{document}